%% file: main.tex
\documentclass[%
 aip,
 amsmath,amssymb,
 reprint,%
]{revtex4-1}

\usepackage{graphicx}
\usepackage{dcolumn}
\usepackage{bm}

\usepackage[utf8]{inputenc}
\usepackage[T1]{fontenc}
\usepackage{mathptmx}
\usepackage{etoolbox}

\makeatletter
\def\@email#1#2{%
 \endgroup
 \patchcmd{\titleblock@produce}
  {\frontmatter@RRAPformat}
  {\frontmatter@RRAPformat{\produce@RRAP{*#1\href{mailto:#2}{#2}}}\frontmatter@RRAPformat}
  {}{}
}%
\makeatother
\begin{document}

\preprint{AIP/123-QED}

\title[ ]{Sensor optimization for urban wind estimation with cluster-based probabilistic framework}
\author{Yutong Liang}
\author{Chang Hou}
\affiliation{School of Robotics and Advanced Manufacture, Harbin Institute of Technology, Shenzhen 518055, Peoples' Republic of China}

\author{Guy Y.~Cornejo Maceda}
\author{Andrea Ianiro}
\author{Stefano Discetti}
 \affiliation{Department of Aerospace Engineering, Universidad Carlos III de Madrid, Av.~de la Universidad, 30, Legan{é}s, 28911, Madrid, Spain}

\author{Andrea Meil{á}n-Vila}
 \affiliation{Department of Statistics, Universidad Carlos III de Madrid, Av.~de la Universidad, 30, Legan{é}s, 28911, Madrid, Spain}

\author{Didier Sornette} \author{Sandro Claudio Lera}
 \affiliation{Institute of Risk Analysis, Prediction and Management, Southern University of Science and Technology, Shenzhen 518055, People's Republic of China}

\author{Jialong Chen} 
\affiliation{Meituan Technology Co., Ltd, Shenzhen 518131, People’s Republic of China}

\author{Xiaozhou He}%
\affiliation{School of Robotics and Advanced Manufacture, Harbin Institute of Technology, Shenzhen 518055, Peoples' Republic of China}
\email{hexiaozhou@hit.edu.cn}

\author{Bernd R.~Noack}%
\affiliation{School of Mechatronics and Control Engineering, Shenzhen University, Canghai campus, Shenzhen 518060, Peoples' Republic of China}
\affiliation{Guangdong Province VTOL Aircraft Manufacturing Innovation Center, Shenzhen 518060,  People’s Republic of China}
\affiliation{School of Robotics and Advanced Manufacture, Harbin Institute of Technology, Shenzhen 518055, Peoples' Republic of China}
\affiliation{Department of Aerospace Engineering, Universidad Carlos III de Madrid, Av.~de la Universidad, 30, Legan{é}s, 28911, Madrid, Spain}
\email{bernd.noack@szu.edu.cn}

\date{\today}

\begin{abstract}
We propose a physics-informed machine-learned framework
for sensor-based flow estimation 
for drone trajectories in complex urban terrain.
The input is a rich set of flow simulations 
at many wind conditions.
The outputs are velocity and uncertainty estimates
for a target domain and subsequent sensor optimization for minimal uncertainty.
The framework has three innovations
compared to traditional flow estimators.
First, the algorithm scales proportionally to the domain complexity, making it suitable for flows that are too complex
for any monolithic reduced-order representation.
Second, the framework extrapolates 
beyond the training data, e.g., smaller and larger wind velocities.
Last, and perhaps most importantly, the sensor location is a free input,
significantly extending the vast majority of the literature.
The key enablers are 
(1) a Reynolds number-based scaling of the flow variables,
(2) a physics-based domain decomposition,
(3) a cluster-based flow representation for each subdomain, (4) an information entropy correlating the subdomains, 
and (5) a multi-variate probability function relating sensor input and targeted velocity estimates.
This framework is demonstrated using drone flight paths
through a three-building cluster as a simple example. We anticipate adaptations and applications for estimating complete cities and incorporating weather input.
\end{abstract}

\maketitle

\input{S1_Intro}

\input{S2_Configuration}

\input{S3_Model}
\input{S4_Results}

\input{S5_Conclusions}

\appendix

\input{S6_App_A}

\input{S6_App_B}
\input{S6_App_C}

\input{S6_App_D}
\input{S6_App_E}

\begin{acknowledgments}
This work is supported by the Shenzhen Science and Technology Program under grants KJZD20230923115210021, JCYJ20220531095605012, JCYJ20240813104853070 and GXWD20220818113020001, 
by the National Science Foundation of China (NSFC) through grants 12172109, 12302293, and 12372216,
and  by the project EXCALIBUR (Grant No PID2022-138314NB-I00), funded by MCIU/AEI/ 10.13039/501100011033 and by ``ERDF A way of making Europe'', and by the funding under ``Orden 3789/2022, del Vicepresidente, Consejero de Educaci\'on y Universidades, por la que se convocan ayudas para la contrataci\'on de personal investigador predoctoral en formaci\'on para el a$\rm{\tilde{n}}$o 2022''. D.S. and S.L. are partially supported by the National Natural Science Foundation of China (Grant No. T2350710802 and No. U2039202), Shenzhen Science and Technology Innovation Commission Project (Grants No. GJHZ20210705141805017 and No. K23405006), and the Center for Computational Science and Engineering at Southern University of Science and Technology.

In addition, we appreciate valuable discussions 
with H.~Li, F.~Raps, J.~Yang, and  Y.~Yang. 
\end{acknowledgments}

\section*{Data Availability Statement}
The data that support the findings of this study are available from the corresponding 
author upon reasonable request.

\nocite{*}
\bibliography{main}

\end{document}

%% file: S1_Intro.tex
\section{Introduction}
\begin{figure*}[htb]
\includegraphics[scale=1]{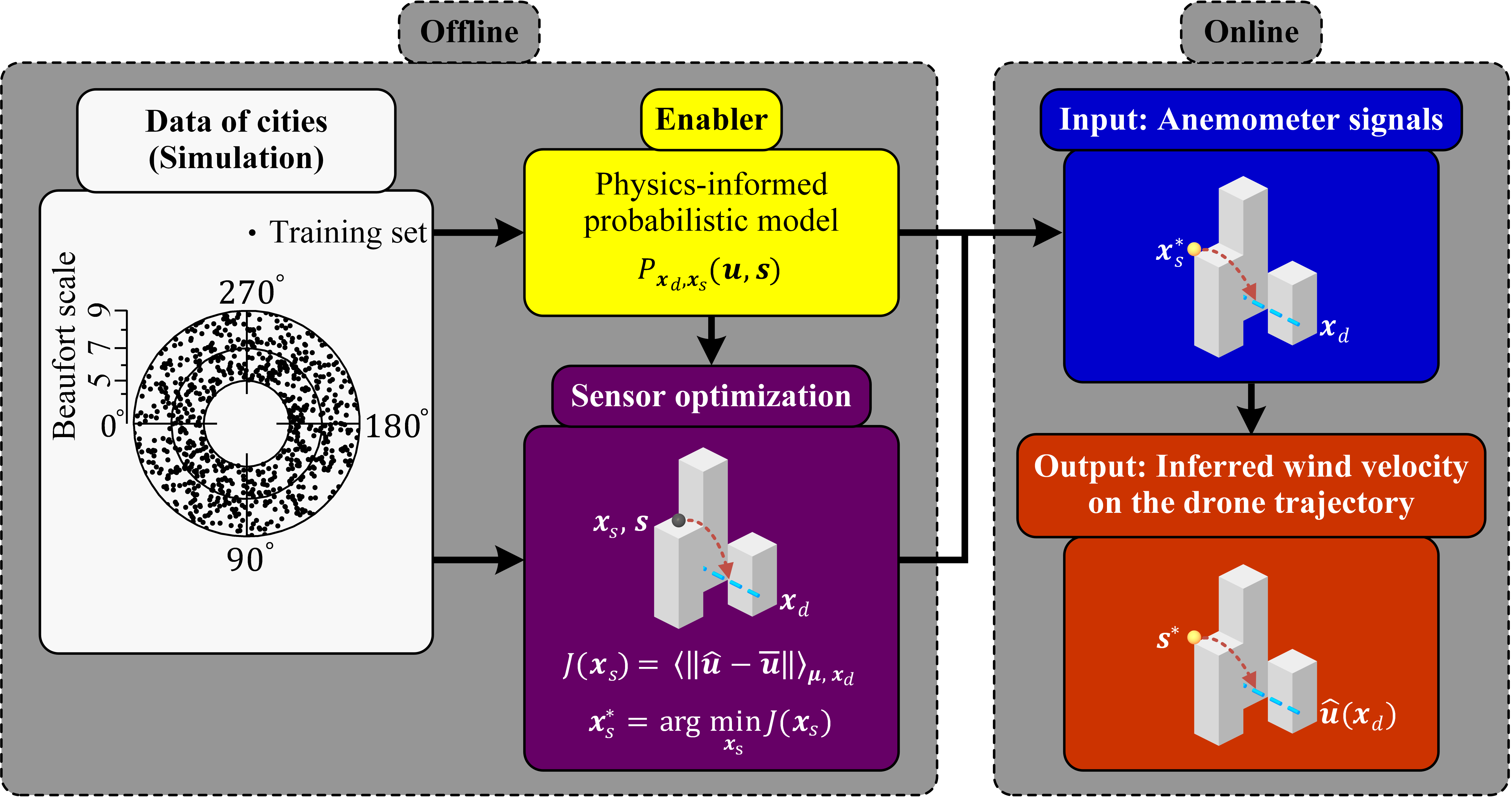}
\caption{\label{Frame} Methodology framework. 
In the offline stage, the simulation dataset of the building complex is generated.
The dataset contains random wind speeds ranging from $7.9$ m/s to $20.7$ m/s (Beaufort wind levels 5 to 8), and incoming flow directions randomly distributed between $0^\circ$ and $360^\circ$.
A physics-informed probabilistic model is built from the training data.
First, the entire flow field is divided into several subdomains, and correlations between these subdomains are established by applying clustering within coarse-grained regions.
Flow estimation is performed by inferring the correlations between the subdomains.
The optimized sensor locations $\bm{x}^*_{s}$ are identified by selecting the most informative sensors.
In the online stage, the velocity field along the drone trajectory is estimated by the sensor signal $\bm{s}^*$ from the optimized sensor locations $\bm{x}^*_s$.}
\end{figure*}
Flow estimation in complex fluid systems 
is inherently challenging due to high dimensionality, 
nonlinearity, and limited or noisy sensor signal data~\cite{gao2024urban}.
Accurate and efficient estimation techniques 
are essential for real-time monitoring, control, and prediction 
in both fundamental and applied fluid dynamics.
Reduced-order modeling (ROM) enables efficient and tractable flow description 
by extracting the dominant flow features~\cite{brunton2020machine},
thereby offering an efficient data-driven approach for estimating large-scale flow dynamics 
while retaining low computational cost and strong physical interpretability.
Among various ROM techniques, cluster-based reduced-order models (CROMs)~\cite{Burkardt2006siam,kaiser2014cluster} have gained increasing attention as a data-driven alternative to classical projection-based methods such as proper orthogonal decomposition (POD)~\cite{Holmes2012book}.

The low computational cost of cluster-based analysis was initially demonstrated in incompressible flow applications~\cite{burkardt2006centroidal}.
Ref.~\onlinecite{kaiser2014cluster} formalized the Cluster-based Markov Model (CMM), modeling the temporal evolution of flow fields as a Markov process over discrete clusters.
Ref.~\onlinecite{fernex2021cluster,li2021cluster} extended this framework by introducing a network-based approach that enables automated construction of reduced-order models from time-resolved data.
Subsequent advancements have led to variations of cluster-based network models capable of capturing nonlinear dynamics, multi-attractor structures, and multi-frequency behaviors, with a focus on automation and robustness~\cite{deng2022cluster,hou2024dynamics}.

%
Urban wind field estimation, characterized by complex multiscale flows around buildings~\cite{teng2025atmospheric}, stands to benefit from the advances of CROMs.
Recent works integrating experiments, simulations, and data-driven models have enhanced both the accuracy and efficiency of urban wind predictions~\cite{sousa2018improving,raissi2020hidden,raissi2019physics,haghighat2021physics,qin2025modeling}.
As the CROM-based framework significantly improves physical interpretability while providing robustness and flexibility for handling multiple flow conditions, it can be expected to be well suited for urban wind estimation. 
This is critical for several tasks, such as trajectory planning of aerial vehicles in urban environments.

Although data-driven models have been applied to estimate velocities along drone trajectories, 
they frequently encounter scalability limitations. 
As the number of spatial query points increases, 
the computational cost of inferring flow fields from sensor data rises sharply. 
Consequently, current methods face challenges in 
balancing accuracy, uncertainty, and model complexity, 
which constrains their industrial applicability.

As shown in Fig.~\ref{Frame}, 
we propose a physics-informed machine-learned framework 
for sensor-based drone trajectory flow estimation.
The key enabler of the framework is a probabilistic model, 
which uses sensor signals 
to estimate the velocity of drone trajectories
based on the drone position and sensor location.
The proposed framework maximizes the accuracy of drone trajectory velocity estimation while mitigating the exponential increase in computational cost associated with growing numbers of sensors and query points.
%

%% file: S2_Configuration.tex
\section{Configuration}
\label{sec2}
\subsection{Numerical simulation}
\begin{figure}[htb]
\includegraphics[scale=1]{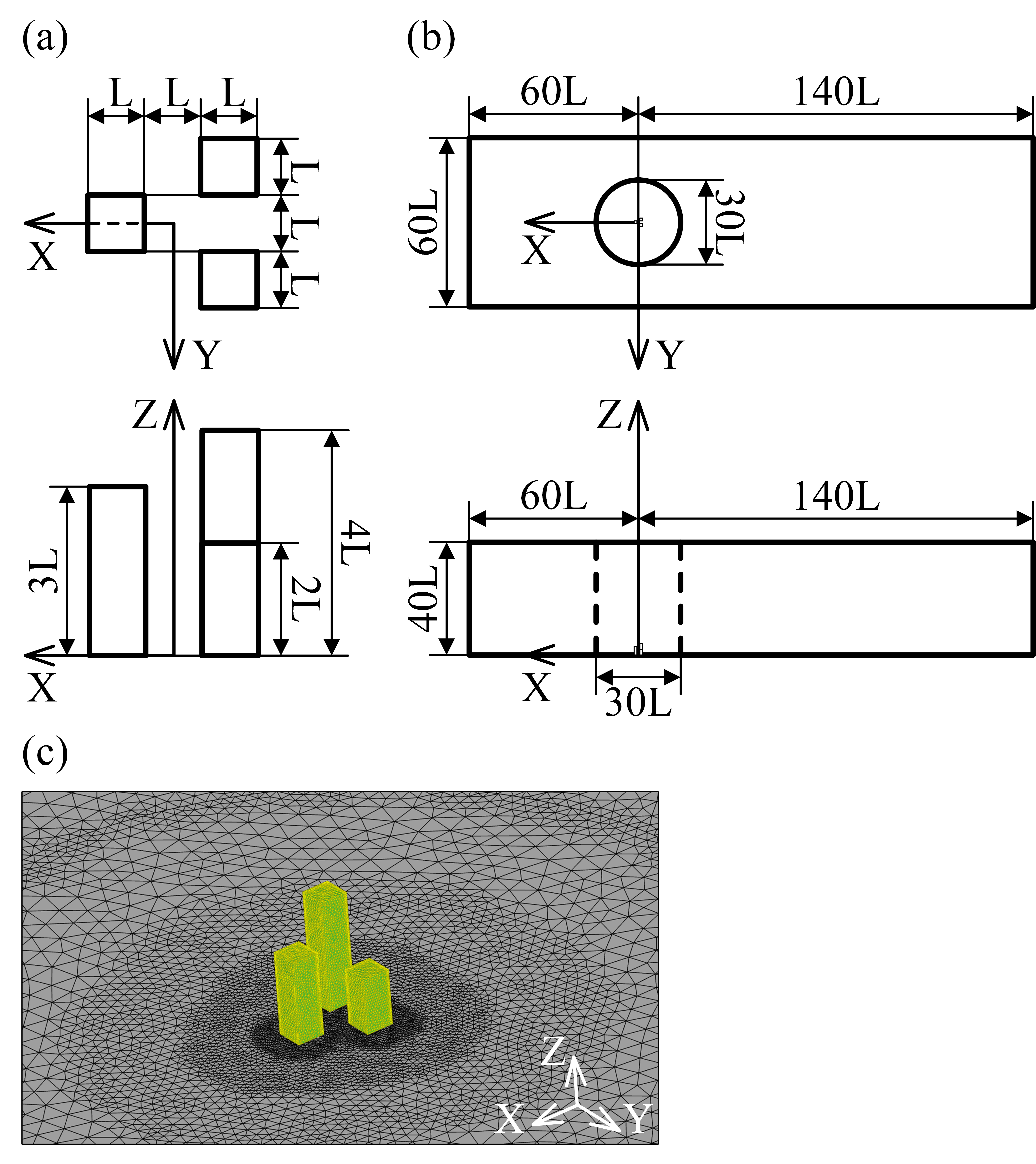}
\caption{\label{BS} Sketch of the building complex.
(a) The top view of the building complex in the $xy$-plane, and the side view in the $xz$-plane. 
The length used for normalization is $L=0.5$~m.
The buildings within the complex are numbered from tallest to shortest as 1, 2, and 3.
(b) The top view of the computational domains in the $XY$ plane, and the side view in the $xz$-plane.
The entire computational domain consists of the inner domain and the outer domain.
(c) Computational grid around the building complex.
}
\end{figure}
To empirically demonstrate our method, we have examined the building complex \cite{liu2025aerodynamic,wang2024coarse}.
Taking the origin at the center of the complex, 
the entire domain is represented using a Cartesian coordinate system.
As shown in Fig.~\ref{BS}(a), each building features a square cross-section with dimensions $L \times L$ (with $L=0.5$~m).
The buildings are labeled 1, 2, and 3 from tallest to shortest, 
with heights of $4L$, $3L$, and $2L$, respectively, where $L = 0.5$~m. 
The projection centers of buildings 1, 2, and 3 in the $xy$-plane 
are located at $(-L, -L)$, $(L, 0)$, and $(-L, L)$, respectively.

We simulate the wind flow around the building complex by solving 
the non-dimensional incompressible Reynolds-averaged Navier–Stokes (RANS) equations. 
As shown in Fig.~\ref{BS}(b), 
the computational domain is partitioned into an inner domain and an outer domain. 
The outer domain extends $200L$, $60L$, and $40L$ in the $x$-, $y$-, and $z$-directions, respectively. 
The cylindrical inner domain has a diameter of $30L$ and a height of $40L$ in the $z$-direction. 
It is designed to enable changes in the incoming wind angle by rotating the domain. 
The inlet and outlet are located $60L$ and $140L$ from the origin, respectively. 
At the inlet, a uniform streamwise velocity is imposed. 
At the outlet, a Neumann condition for velocity and a Dirichlet condition for pressure are applied. 
A no-slip condition is enforced on the surfaces of the building complex and the ground. 
Interface conditions are imposed at the junction between the inner and outer domains. 
Slip conditions are applied on the remaining boundaries to prevent wake–wall interactions. 
Figure~\ref{BS}(c) shows a magnified view of the grid around the building complex.

The training set $\mathcal{D}_{\rm train}$ and the testing set $\mathcal{D}_{\rm test}$ consist of 800 and 200 snapshots, respectively. 
The wind velocity magnitude $U_\infty$ in the dataset is randomly sampled between 7.9 and 20.7~m/s (corresponding to Beaufort levels 5–8), 
covering a broad range of realistic wind conditions typically encountered in operational environments. 
The wind direction $\alpha$ is randomly sampled from $0^\circ$ to $360^\circ$, 
ensuring representation of all possible inflow angles. 
The dataset is designed to incorporate diverse and realistic inflow conditions, 
thereby enhancing the model’s generalization ability and providing a closer approximation to actual atmospheric variability in urban and complex terrain environments.

\subsection{Verification}
\begin{table}
\caption{\label{tab:grid} 
Grid information for the independence test.}
\begin{ruledtabular}
\begin{tabular}{lccc}
\multicolumn{1}{l}{Grids} & \multicolumn{1}{c}{Inner domain} & \multicolumn{1}{c}{Outer domain} & \multicolumn{1}{c}{Number of points} \\ 
\multicolumn{1}{c}{ } & \multicolumn{1}{c}{($\times 10^4$)} & \multicolumn{1}{c}{($\times 10^4$)} & \multicolumn{1}{c}{($\times 10^4$)} \\ 
\hline
1 & 22 & 108 & 23\\
2 & 47 & 108 & 27\\
3 & 92 & 108 & 35\\
4 & 210 & 108 & 55\\
5 & 283 & 108 & 68\\
6 & 416 & 108 & 90\\
7 & 517 & 108 & 108\\
8 & 678 & 108 & 136\\ 
\end{tabular}
\end{ruledtabular}
\end{table}
\begin{figure}[htb!]
\includegraphics[scale=1]{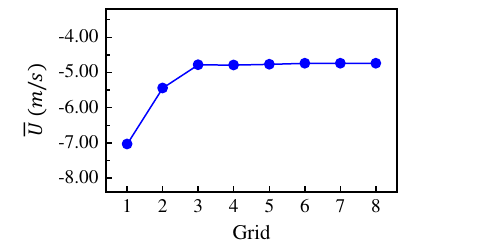}
\caption{\label{GI} Results of the grid independence test conducted using 8 grid sets.
The plot shows the mean streamwise velocity $\bar{U}$ across a $4L$ line parallel to the $z$-axis at location ($x=1.75L, y=0$).
}
\end{figure}
Before extensive simulations, 
grid independence tests were performed at a wind velocity of 2 m/s and a wind angle of $0^\circ$ to determine the optimal grid resolution. 
Eight different configurations with different densities were tested in the inner domain, 
as summarized in Table~\ref{tab:grid}. 
The variation of the mean streamwise velocity $\bar{U}$ 
along the line at $(x=1.75L, y=0)$ with different grid densities is shown in Fig.~\ref{GI}.
Based on these results, Grid configuration~5 was selected for subsequent simulations.

%% file: S3_Model.tex
\section{Cluster-based Probabilistic Framework}
\label{sec3}
\begin{figure*}[t]
\includegraphics[scale=1]{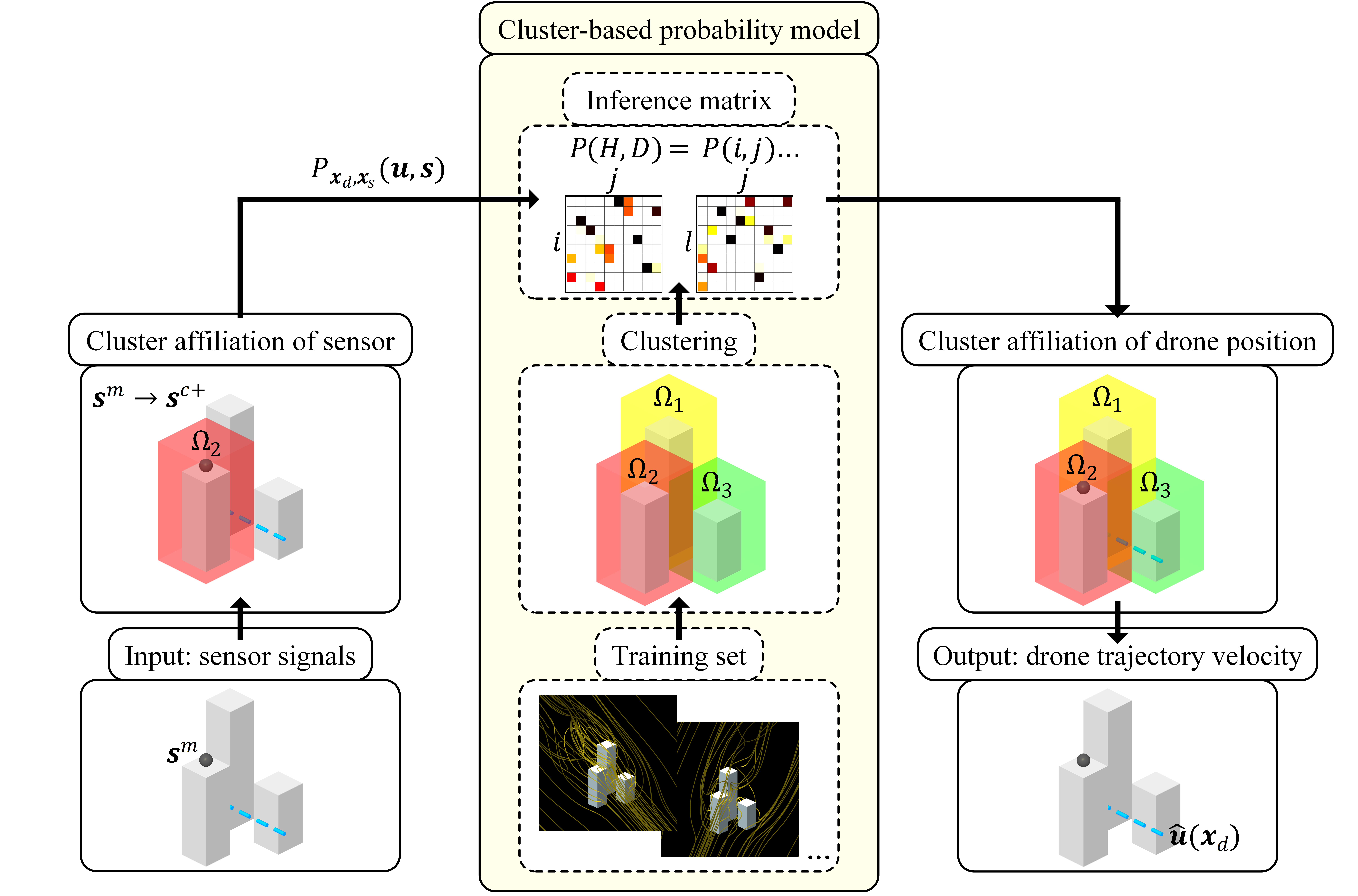}
\caption{\label{Model} Sensor-based flow estimation
exemplified for a single sensor.
The training sensor signal $\bm{s}^m$ serves as input, from which the sensor’s cluster affiliation $\bm{c}_{1,i}(\bm{s})$ is inferred.
This cluster affiliation $\bm{c}_{1,i}(\bm{s})$ is the input to the physics-informed probabilistic model, which then infers the cluster affiliation $\bm{c}_{2,j}(\bm{x}_d)$ of the subdomain containing the drone trajectory $\bm{x}_d$, ultimately estimating the velocity along the drone trajectory.
The dashed box indicates the physics-informed probabilistic model, constructed by first decomposing the training flow field snapshots $\bm{u}^m$ into subdomains $\Omega_1, \Omega_2, \Omega_3$, then clustering each subdomain, and finally establishing the inference matrix $\mathrm{P}$ between them.
}
\end{figure*}
The cluster-based probabilistic framework 
consists of offline and online steps as shown in Fig.~\ref{Frame}.
\subsection{Non-dimensionalization of the data}
The training set $\cal{D}_{\rm train}$ calibrates the probabilistic model for sensor optimization.
The training set $\cal{D}_{\rm train}$ and the testing set $\cal{D}_{\rm test}$ are defined as
\begin{equation}
    {\cal{D_{\rm train}}} := \left\{ \bm{\mu}^m, \bm{u}^m (\bm{x} ) \right\}_{m=1}^M,
\end{equation}
\begin{equation}
    {\cal{D}}_{\rm test} := \left\{ \bm{\mu}^{M+n}, \bm{u}^{M+n} (\bm{x} ) \right\}_{n=1}^N,
\end{equation}
where $\bm{u}^m(\bm{x})$ denotes the $m$-th snapshot at location $\bm{x}$. Calibration and testing are performed for $m \in \{1,\ldots,M\}$ and $n \in \{M+1,\ldots,M+N\}$, respectively.
As shown in Fig.~\ref{Frame}, a sufficient range of operating conditions is covered by the data set,
for each snapshot, the operating parameters $\bm{\mu}^m$ are random values varying within a certain range.
The corresponding sensor input for the training set and testing set, denoted as ${\cal{S}}_{\rm{train}}$ and ${\cal{S}}_{\rm{test}}$ are defined as
\begin{equation}
{\cal{S}}_{\rm{train}} := \left\{ \bm{\mu}^m, \bm{s}^m (\bm{x}_s) \right\}_{m=1}^M,
\end{equation}
\begin{equation}
{\cal{S}}_{\rm{test}} := \left\{ \bm{\mu}^{M+n}, \bm{s}^{M+n}(\bm{x}_s) \right\}_{n=1}^N,
\end{equation}
where $\bm{s}^m(\bm{x}_s)$ 
represents the sensor signal 
at location $\bm{x}_s$ for the $m$-th snapshot.
This article uses sensor signals $\bm{s}$ to illustrate wind velocity at sensor locations $\bm{x}_s$.

All velocity and sensor data are non-dimensionalized with respect to the oncoming wind velocity
\begin{equation}
\bm{u}^{+} := \frac{ \bm{u} } {U_\infty}, \quad
\bm{s}^{+} := \frac{ \bm{s} } {{U}_\infty},
\end{equation}
where $\bm{u}^{+}$ and $\bm{s}^{+}$ denote the normalized flow field velocity and sensor signals, respectively.

\subsection{Cluster-based probabilistic model}
The cluster-based probabilistic model is highlighted by the yellow box in Fig.~\ref{Model}.
\subsubsection{Domain decomposition}
\begin{figure}[hbt!]
\includegraphics[scale=1]{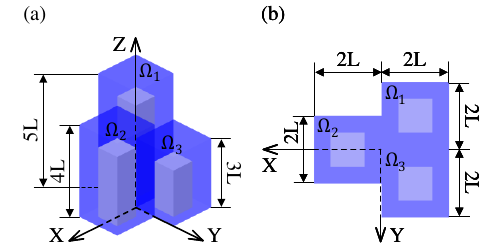}
\caption{\label{DD} Domain decomposition.
(a) The oblique view of the three subdomains.
For each snapshot, 
the entire flow field is decomposed into three subdomains 
$\Omega_1$, $\Omega_2$, and $\Omega_3$ around buildings 1, 2, and 3.
From the tallest to the shortest, 
the heights of the subdomains are $5L$, $4L$ and $3L$,
with $L=0.5$~m.
(b) The top view of the three subdomains.
Each subdomain has a square cross-section measuring $L \times L$.
}
\end{figure}

As illustrated in Fig.~\ref{Frame}, the proposed physics-informed probabilistic framework leverages the training set $\mathcal{D_{\rm train}}$ to perform subdomain-based flow estimation.
Directly modeling the entire urban flow field as a whole is prohibitively complex due to the intricate geometry of the building clusters. 
However, when the flow is narrowed to individual buildings, the associated wakes exhibit strong structural correlations, and the spatial complexity is significantly reduced. 

Therefore, as shown in Fig.~\ref{DD}, 
the normalized training snapshot $\bm{u}^{m+}$ is decomposed into $3$ subdomains, 
denoted as $\Omega_1$, $\Omega_2$,  and $\Omega_3$
corresponding to building $1$, $2$, and $3$ respectively. 
Each subdomain has a square cross-section measuring $2L \times 2L$.
From the tallest to the shortest, 
the heights of the subdomains are $5L$, $4L$, and $3L$.
\subsubsection{Clustering}
In each subdomain, the training data is coarse-grained in $K$ clusters.
The cluster-affiliation function maps the local flow state of the training dataset 
$ \bm{u}^m \in {\cal{D}}_{\rm train}$ to the index of its nearest centroid in each subdomain, 
\begin{equation}
\begin{aligned}
k_1(m) &:= \arg \min _{i} \left\| \bm{u}^{m+} - \boldsymbol{c}_{1,i} \right\|_{\Omega_1}, \\
k_2(m) &:= \arg \min _{j} \left\| \bm{u}^{m+} -  \boldsymbol{c}_{2,j} \right\|_{\Omega_2}, \\
k_3(m) &:= \arg \min _{l} \left\| \bm{u}^{m+}  - \boldsymbol{c}_{3,l} \right\|_{\Omega_3}.
\end{aligned}
\end{equation}
Here,  $\|\cdot\|_{\Omega}$ denotes the Hilbert norm over each subdomain, and $\bm{c}_{1,i}$, $\bm{c}_{2,j}$, and $\bm{c}_{3,l}$ are the centroids of the $i$-th, $j$-th, and $l$-th clusters in subdomains 1, 2, and 3, respectively, for $i,j,l = 1, \dots, K$.

\subsubsection{Inference matrix}
The spatial correlations between subdomains are captured by the inference matrix $\mathbf{P}$, 
which encodes the 
conditional probabilities $P_{ji}$ between cluster affiliations in the two considered subdomains. 
For example, the conditional probability that $\Omega_2$ belongs to cluster $j$ given that $\Omega_1$ belongs to cluster $i$ is defined as:
\begin{equation}
P_{ji} := \frac{\sum_{m=1}^M  \delta _{i,k_1(m)} \times \delta _{j,k_2(m)} }
{\sum_{m=1}^M \delta _{i,k_1(m)}},
\end{equation}
where
\begin{equation}
\delta _{ij}=
\left\{\begin{matrix}
 1, & \text{if }i=j,\\
 0, & \text{otherwise}.
\end{matrix}\right.
\end{equation}
The inference matrix $\mathbf{P}=\left (P_{ji}\right)_{i,j}$ enables the inference of the target subdomain state based on the known cluster
affiliation in a reference subdomain, typically where the sensors are located.
The stochasticity of the inference matrix $\mathbf{P}$ can be quantified using the Kullback-Leibler entropy~\cite{kaiser2014cluster,hou2024dynamics}, which characterizes the uncertainty of the inference.

\subsection{Flow estimation from sensor signals}
Given the sensor signal $\bm{s}=\bm{s}^m(\bm{x}_s)$ at different locations $\bm{x}_s$, the inference of the cluster affiliation in the domain of the sensor location utilizes the $k$NN algorithm (see Appendix A).
Assuming, for instance, that the sensor is located at $\Omega_1$, the drone trajectory is located at $\Omega_2$.
Once the source cluster $\bm{c}_{1,i}$ is identified, the inference matrix $\mathbf{P}$ provides the distribution of probable clusters $\bm{c}_{2,j}$ for the target location $\bm{x}_d$. 

The estimated velocity $\hat{\bm{u}}$ at $\bm{x}_d$ is computed by the expectation over the inferred probability:
\begin{equation}
\hat{\bm{u}}(\bm{x}_d, \bm{x}_s, \bm{s} ) 
= 
\int 
\bm{u} \> 
P_{\bm{x}_d, \bm{x}_s} (\bm{u}, \bm{s} )
d\bm{u} \> .
\end{equation}
Here, 
$ P_{\bm{x}_d, \bm{x}_s} (\bm{u}, \bm{s} ) $
characterizes the uncertainty when inferring the velocity field $\bm{u}$ at drone trajectory location $\bm{x}_d$,
given the sensor signal $\bm{s}$ at sensor location $\bm{x}_s$.
$P_{\bm{x}_d, \bm{x}_s} (\bm{u}, \bm{s} ) $ is obtained with $P_{ji}$ from the inference matrix $\mathbf{P}$.

The drone trajectory for wind estimation 
is parametrized by
\begin{equation}
\beta \in [0,1]  \mapsto \bm{x}_d[\beta].
\end{equation}
The estimation error for snapshot $m$ is defined by 
\begin{equation}
{E}^m :=   
\int_0^1 \! 
\Vert\hat{\bm{u}}^m\left(\bm{x}_d[\beta]\right) - \bm{u}^m\left(\bm{x}_d[\beta]\right) \Vert^2
d\beta \>,
\end{equation}
here, $\hat{\bm{u}}^m$ is the velocity estimated at $\bm{x}_d$ using the sensor signal $\bm{s}^m$, while $\bm{u}^m$ denotes the corresponding ground-truth mean flow field velocity from $\mathcal{D}_{\rm{train}}$. 
The case error $E^m$ quantifies the discrepancy between estimated and ground-truth velocities along the trajectory at the $m$-th training case.
The average estimation error over the training set $\mathcal{D}_{\rm{train}}$ is defined by
\begin{equation}
{E}({\cal{D}}_{\rm train}) :=  \frac{1}{M} 
\sum_{m = 1} ^{M} {E} ^ m,
\end{equation}
The average estimation error $E(\mathcal{D}_{\rm{train}})$ quantifies the average discrepancy between estimated and ground-truth velocities along the trajectory across the entire training set.

\subsection{Sensor optimization}
After constructing the physics-informed probabilistic model, the subsequent sensor optimization process leverages the estimated flow field to improve the placement strategy. 
The optimal sensor location $\bm{x}^*_s$ is then defined as the configuration that minimizes the average estimation error:
\begin{equation}
\bm{x}^*_s := \arg \min_{\bm{x}_s} {E}({\cal{D}}_{\rm{train}}),
\end{equation}
where $\bm{x}_s^*$ represents the sensor location that yields the lowest average estimation error ${E}({\cal{D}}_{\rm{train}})$. 



%% file: S4_Results.tex
\section{Results}
\label{sec4}
\begin{figure*}[htb]
\includegraphics[scale=1]{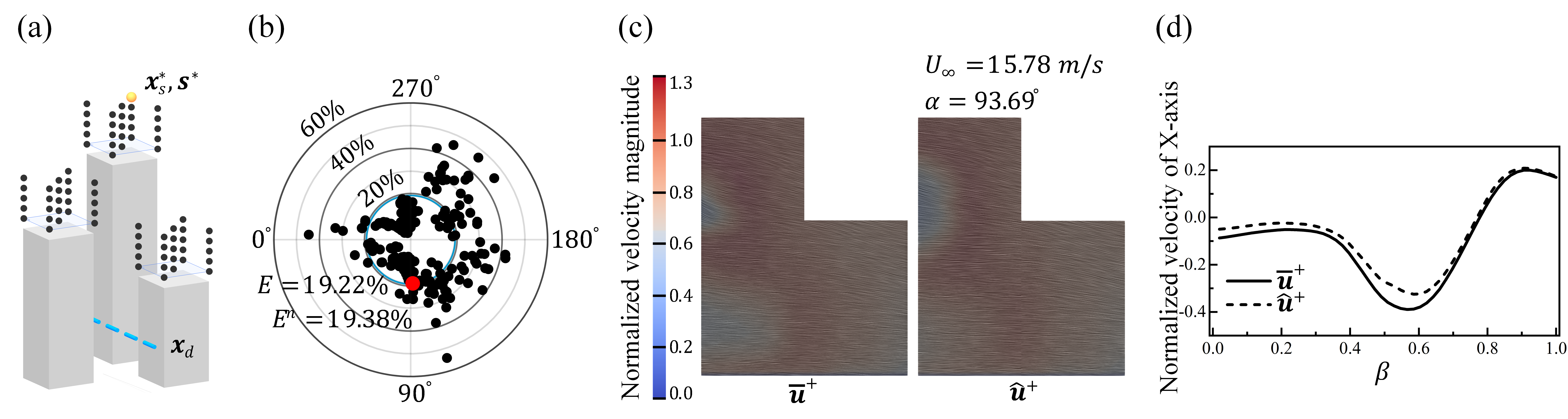} 
\caption{\label{Res}
Flow estimation results for the building complex.
(a) Sensor location candidates: black dots denote original sensor positions, the yellow dot is the optimized sensor $\bm{x}^*_s$, and the blue dashed line indicates the drone trajectory $\bm{x}_d$. 
Axes correspond to the representative case 
where the error  for the optimized sensor $E^n$ approximates its average error $E({\cal{D}}_{\rm{test}})$.
(b) Estimation error $E^{M+n}$ vs.\ wind direction $\alpha$: 
blue circle shows the average estimation error of $\bm{x}^*_s$;
black dots represent estimation errors $E^{M+n}$ with respect to $\alpha$;
the red dot marks the case where $E^{M+n} \approx E({\cal{D}}_{\rm{test}})$.
(c) Representative slice at $X=0$ for the marked case:
$\overline{\bm{u}}$ is the reference from ${\cal{D}}_{\rm{test}}$, and $\hat{\bm{u}}$ is the estimated field using $\bm{x}^*_s$.
(d) Mean and estimated velocities along $X$ on the trajectory of the marked case.
}
\end{figure*}
The proposed framework is demonstrated on the building complex dataset under varying wind conditions, see Fig.~\ref{Frame}.
Sensor optimization is first performed to identify the optimal sensor location $\bm{x}_s^*$ that minimizes the average estimation error $E(\mathcal{D}_{\rm train})$, 
followed by evaluation on the testing set.
Model parameters are detailed in Appendix~B.

As shown in Fig.~\ref{Res}(a), 
there are 25 candidate sensor locations on each building. 
The optimized sensor is positioned at $(-1.5L,-1.5L,5L)$ above the tallest building, 
and the drone trajectory spans the $xz$-plane at constant height $L$.
The selected wind condition corresponds to a case 
where the estimation error $E^{M+n}$ is close to the average testing error $E(\mathcal{D}_{\rm test})$.
For this sensor configuration, $E(\mathcal{D}_{\rm test})=19.22\%$, the chosen case error is $19.38\%$.
Nearly $60\%$ of case errors fall below $20\%$, about $90\%$ are under $30\%$, and fewer than $5\%$ exceed $40\%$, demonstrating the model's robustness and accuracy.
Figure~\ref{Res}(c) compares the mean velocity $\overline{\bm{u}}$ from the testing set and the estimated velocity $\hat{\bm{u}}$ for the selected case.
Despite $E^{M+n} \approx 20\%$, the estimated field closely reproduces both the magnitude and distribution of the reference, indicating reliable directional and intensity inference.
The velocity profiles along the trajectory are shown in Fig.~\ref{Res}(d). 
Collisions between drones and buildings might be caused by huge estimation error in the $x$-direction wind flow field. 
Since this component is more critical for drone operation, 
we focus exclusively on the $x$-direction in the present analysis.
With the estimation error $E^{M+n}=19.38\%$,
the estimated velocity $\hat{U}^+$ matches the reference $\overline{U}^+$ well, 
further confirming the estimation fidelity along critical flight paths.

%% file: S5_Conclusions.tex
\section{Conclusions}
\label{sec5}
Summarizing, 
we propose a versatile physics-informed machine-learned framework for sparse sensor-based estimation of complex fields at a large range of operating conditions.
This framework addresses key challenges of sensor-based field estimators related to the extent of the domain, the amount of data 
and the need to optimize sensor positions.
We successfully demonstrate
a sensor optimization for flow estimation on a drone trajectory around a building cluster. 

The innovations are demonstrated 
with respect to a traditional 
sensor-based mapping from signals to wake flows
under different operating conditions \cite{LiSQ2022jfm}.
Evidently, a monolithic reduced-order representation comprising uncorrelated events will lead to excessive dimensions, mitigating the chances for sparse sensing.
Hence, we partition the domain 
and apply a cluster-based approximation to each subdomain.
Correlations between the subdomain states are quantified with the Kullback-Leibler entropy 
from the inference matrix. 
Thus, the computational cost for the offline calibration 
and online estimation scales linearly with the complexity of the flow.
Second, high-Reynolds-number turbulence
features 
independence of non-dimensional quantities 
from the Reynolds number~\cite{HouC2024pf}.
This turbulence property 
 allows for a scaling that extrapolates 
 existing databases and thus dramatically reduces 
 the required data. 
Finally, the proposed probabilistic flow representation enables inferences for arbitrary inquiry points from arbitrary sensor locations. 
Thus, sensor optimization can be performed with a computationally low-cost plant.

Evidently, 
the proposed estimation framework 
can accommodate a large amount of sensor information, 
even weather information
and can be employed for a large range of problems with similar spatio-temporal features.
Future improvements can, for instance, 
be achieved by generalizing the discrete cluster representation with continuous affine cluster-based expansions.

%% file: S6_App_A.tex
\section{Sensor cluster affiliation inference using $k$-nearest neighbours method}
\label{App.kNN}
\begin{figure}[hbt!]
\includegraphics[scale=1]{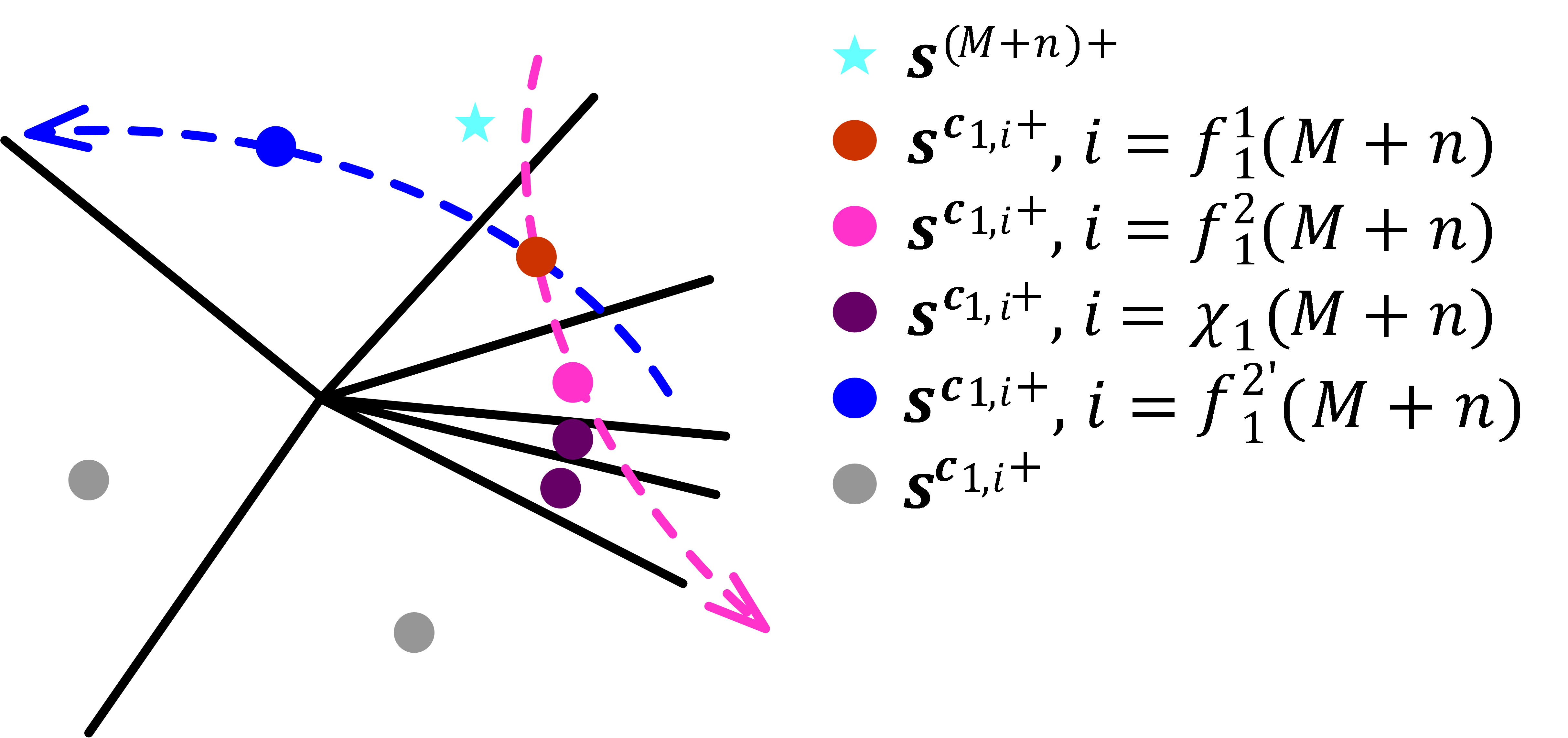}
\caption{\label{kNN}
Sketch of sensor cluster affiliation inference using $k$-nearest neighbours method with $k=2$.
The light blue star denotes the normalized sensor signal $\bm{s}^{(M+n)+}$ 
of the testing set $\mathcal{D}_{\mathrm{test}}$.
The red dot denotes the normalized centroid sensor signal $\bm{s}^{\mathbf{c}_{1,i}^{+}}$ nearest to $\bm{s}^{(M+n)+}$,
using ``first-centroid'' to denote this centroid.
The navy blue dot denotes the normalized centroid sensor signal $\bm{s}^{\mathbf{c}_{1,i}^{+}}$ 
second nearest to $\bm{s}^{(M+n)+}$.
The pink dot and purple dot denote the normalized centroid sensor signals $\bm{s}^{\mathbf{c}_{1,i}^{+}}$ from the three centroids nearest to the first-centroid. 
Among these three sensor signals, 
the normalized centroid sensor signal $\bm{s}^{\mathbf{c}_{1,i}^{+}}$ represented by the pink dot 
is the nearest to $\bm{s}^{(M+n)+}$.
The gray dots represent the rest normalized centroid sensor signals $\bm{s}^{\mathbf{c}_{1,i}^{+}}$.
The navy blue dashed line represents 
the dynamic trajectory of $\bm{s}^{(M+n)+}$ estimated by 
$\bm{s}^{\mathbf{c}_{1,i}^{+}}(i=f^1_1)$ and $\bm{s}^{\mathbf{c}_{1,i}^{+}}(i=f^{2'}_1)$.
The pink dashed line represents 
the dynamic trajectory of $\bm{s}^{(M+n)+}$ estimated by 
$\bm{s}^{\mathbf{c}_{1,i}^{+}}(i=f^1_1)$ and $\bm{s}^{\mathbf{c}_{1,i}^{+}}(i=f^2_1)$.
}
\end{figure}
Taking subdomain $\Omega_1$ as an example.
Given the testing set sensor signal $\bm{s}^{M+n}$ in subdomain $\Omega_1$,
firstly the coming wind velocity $\bar{U}^+$ was estimated using the $k$-nearest neighbours ($k$NN) algorithm with $k=4$. 
Then normalize the testing set sensor signal $\bm{s}^{M+n}$ using the estimated coming wind velocity $\bar{U}^+$.

For the $k$-nearest neighbours ($k$NN) method with $k=2$, 
the index of the first-centroid nearest to the $\bm{s}^{(M+n)+}$ 
is denoted by $f^1_1(M+n)$.
$f^1_1(M+n)$ is determined by comparing the normalized sensor signal $\bm{s}^{(M+n)+}$ with the normalized signals of the cluster centroids $\bm{s}^{\bm{c}+}$ at the same sensor location $\bm{x}_s$: 
\begin{equation}
f^1_1(M+n) := 
\arg \min_i \left\| \bm{s}^{(M+n)+} - \bm{s}^{\bm{c}_{1,i}+} \right\|_{\Omega_1}.
\end{equation}

Similarly, for the sensor signals collected in the subdomain $\Omega_2$ and $\Omega_3$, the $f^1_2(M+n)$ and $f^1_3(M+n)$ are defined as:
\begin{equation}
f^1_2(M+n) := 
\arg \min_j \left\| \bm{s}^{(M+n)+} - \bm{s}^{\bm{c}_{2,j}+} \right\|_{\Omega_2},
\end{equation}
\begin{equation}
f^1_3(M+n) := 
\arg \min_l \left\| \bm{s}^{(M+n)+} - \bm{s}^{\bm{c}_{3,l}+} \right\|_{\Omega_3}.
\end{equation}

Once the index of the first-centroid nearest to the $\bm{s}^{(M+n)+}$ is determined, 
additional clusters are subsequently selected from the neighbors of the first-centroid 
using a $k$-nearest neighbor search with $k=3$:
\begin{equation}
\chi_1(M+n) := 
\arg \min_i \left\| \bm{c}_{1,i} - f^1_1(M+n) \right\|_{\Omega_1},
\end{equation}
\begin{equation}
\chi_2(M+n) := 
\arg \min_j \left\| \bm{c}_{2,j} - f^1_2(M+n) \right\|_{\Omega_2},
\end{equation}
\begin{equation}
\chi_3(M+n) := 
\arg \min_l \left\| \bm{c}_{3,l} - f^1_3(M+n) \right\|_{\Omega_3},
\end{equation}
where $\chi_1$ denotes the indices of these three nearest neighbours. 

The index of the second-centroid $f^2_1(M+n)$ is obtained by comparing $\bm{s}^{\bm{c}_{1,i}+}$ with the $\bm{s}^{(M+n)+}$ restricted to the neighbourhood $\chi_1$ at the same location $\bm{x}_s$,
\begin{equation}
f^2_1(M+n) := 
\arg \min_i \left\| \bm{s}^{(M+n)+} - \bm{s}^{\bm{c}_{1,i}+} \right\|_{\Omega_1,{\chi_1}},
\end{equation}.

Similarly, for the sensor signals collected in the subdomain $\Omega_2$ and $\Omega_3$, the $f^2_2(M+n)$ and $f^2_3(M+n)$ are defined as:
\begin{equation}
f^2_2(M+n) := 
\arg \min_j \left\| \bm{s}^{(M+n)+} - \bm{s}^{\bm{c}_{2,j}+} \right\|_{\Omega_2,{\chi_2}},
\end{equation}
\begin{equation}
f^2_3(M+n) := 
\arg \min_l \left\| \bm{s}^{(M+n)+} - \bm{s}^{\bm{c}_{3,l}+} \right\|_{\Omega_3,{\chi_3}},
\end{equation}

For $k$-nearest neighbours method with $k=1$,
the first centroid is determined 
in the same way with $k=2$.

%% file: S6_App_B.tex
\section{Parameters used}
\label{App:Pare}
A clustering parameter of $K=20$ was applied to each subdomain. 
The incoming wind velocity $\hat{U}_\infty$ is estimated using a $k$-nearest neighbours ($k$NN) algorithm with $k=4$;  
Both the sensor signal cluster affiliation $\bm{c}_{1,i}(\bm{s})$ and the drone trajectory cluster affiliation $\bm{c}_{2,j}(\bm{x}_d)$ are estimated using $K$-nearest neighbours with $K=2$.

%% file: S6_App_C.tex
\section{Sensor optimization}
\label{App:SensorOpt}
\begin{figure}[htb!]
\includegraphics[scale=1]{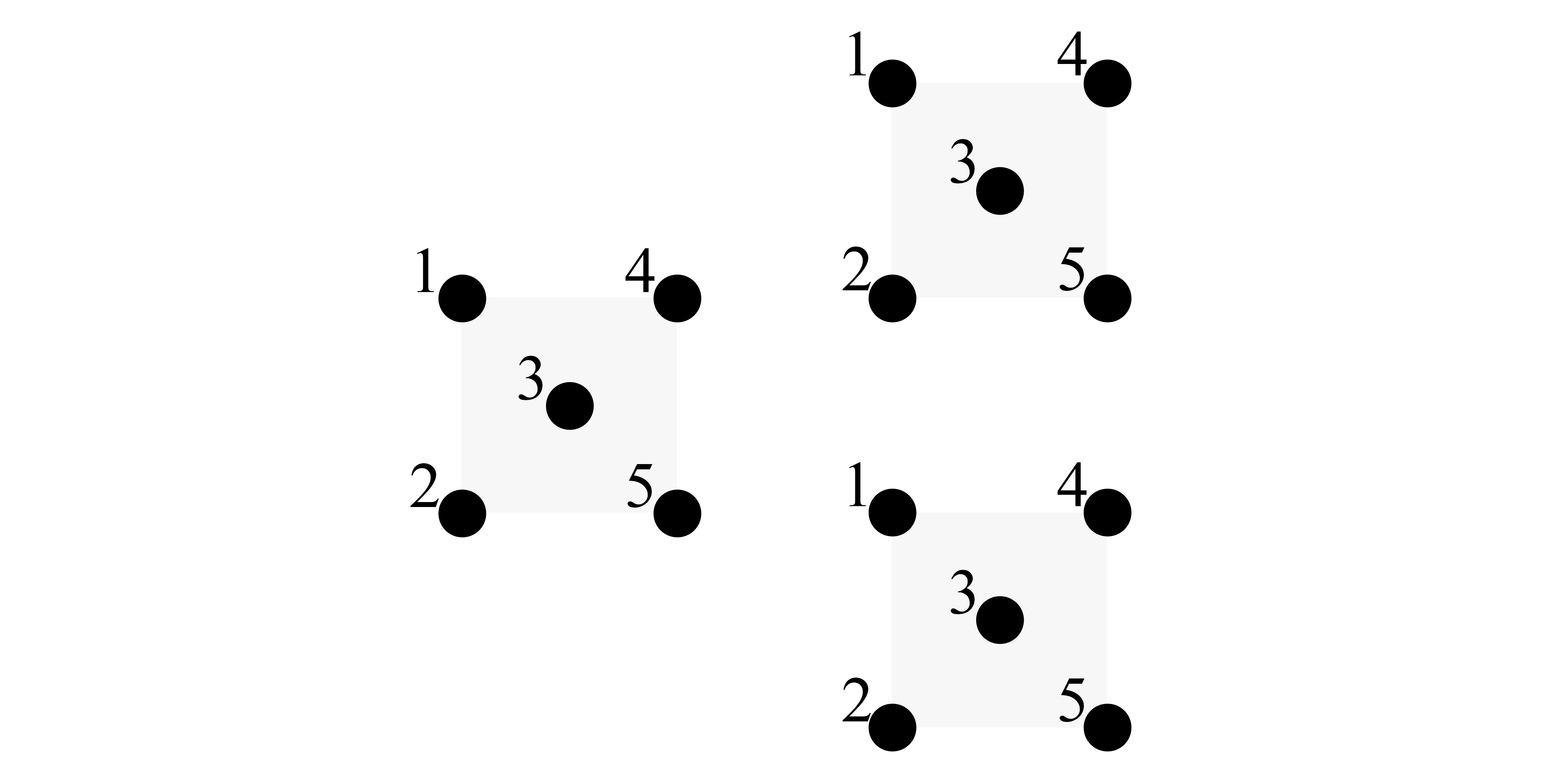}
\caption{\label{App.SL}
Projections of Sensors 1–5 in subdomains $\Omega_1$, $\Omega_2$, and $\Omega_3$ onto the $xy$-plane.
Sensors 1, 2, 4, and 5 project onto the $xy$-plane at the building corners, 
whereas Sensor~3 projects onto the center. 
The $z$-direction heights of Sensors 1–5 are $4.2L$ in subdomain $\Omega_1$, 
$3.2L$ in subdomain $\Omega_2$, and $2.2L$ in subdomain $\Omega_3$.
}
\end{figure}
\begin{figure}[htb!]
\includegraphics[scale=1]{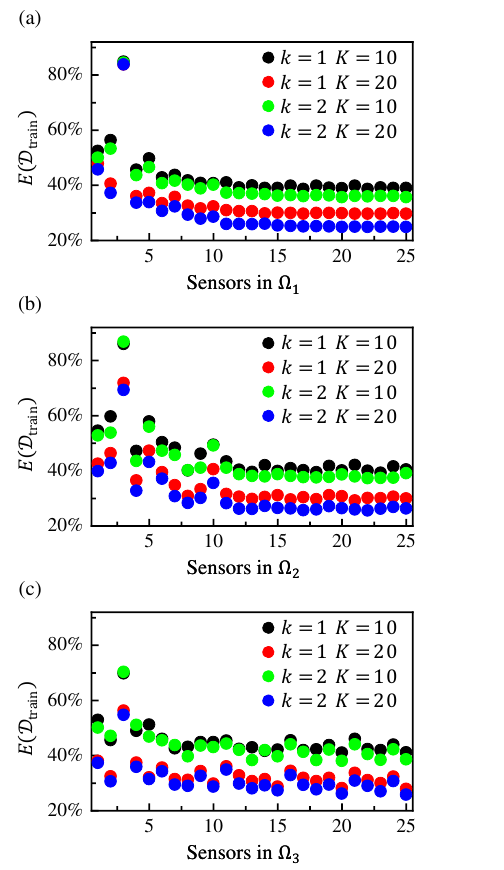}
\caption{\label{App.avgE}Average estimation error $E$ over training set $\mathcal{D}_{\mathrm{train}}$ of the sensors in subdomain $\Omega_1$, $\Omega_2$, and $\Omega_3$.}
\end{figure}
Projections of Sensors 1–5 in subdomains $\Omega_1$, $\Omega_2$, 
and $\Omega_3$ onto the $xy$-plane are shown in Fig.~\ref{App.SL}.
Sensors 6–10 follow the same projection pattern 
on the $xy$-plane as Sensors 1–5, 
but their heights in the $z$-direction are increased by 0.1$L$.
Sensors 11–25 are arranged in the same manner.

The average estimation errors $E$ over 
the training set $\mathcal{D}_{\mathrm{train}}$ for all sensor positions are shown in Fig.~\ref{App.avgE}.
Black dots denote results from the $k$-nearest neighbors method with $k=1$ 
when inferring sensor cluster affiliation using $K=10$ clusters, 
whereas red dots denote the case with $k=1$ and $K=20$, 
green dots denote the case with $k=2$ and $K=10$,
blue dots denote the case with $k=2$ and $K=20$. 
The details of the $k$-nearest neighbors method 
with $k=1$ and $k=2$ can be seen in Appendix~\ref{App.kNN}.
As shown in Fig.~\ref{App.avgE}, the average estimation error $E(\mathcal{D}_{\mathrm{train}})$
at a given sensor location is minimized when $k=2$ and $K=20$. 
Sensor~24 in $\Omega_1$ exhibits the lowest average estimation error $E(\mathcal{D}_{\mathrm{train}})$,
thus Sensor~24 in $\Omega_1$ is the optimal sensor location.

%% file: S6_App_D.tex
\section{Error sources of the framework}
\label{App:ErrorSoures}
\subsection{Representation error}
\label{RepresentationE}
\begin{figure}[hbt!]
\includegraphics[scale=1]{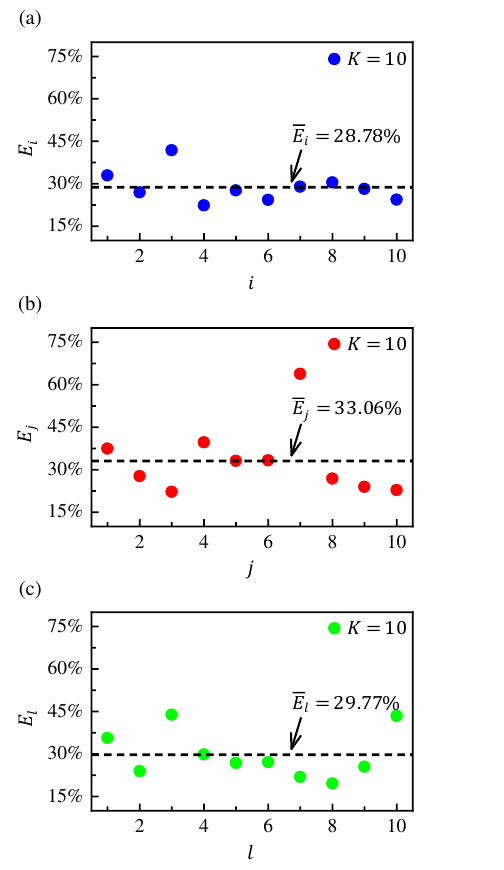}
\caption{\label{App.RE10} Representation errors and the average representation errors for the flow field in subdomains $\Omega_1$, $\Omega_2$, and $\Omega_3$ with clustering number $K=10$.
}
\end{figure}
\begin{figure}[hbt!]
\includegraphics[scale=1]{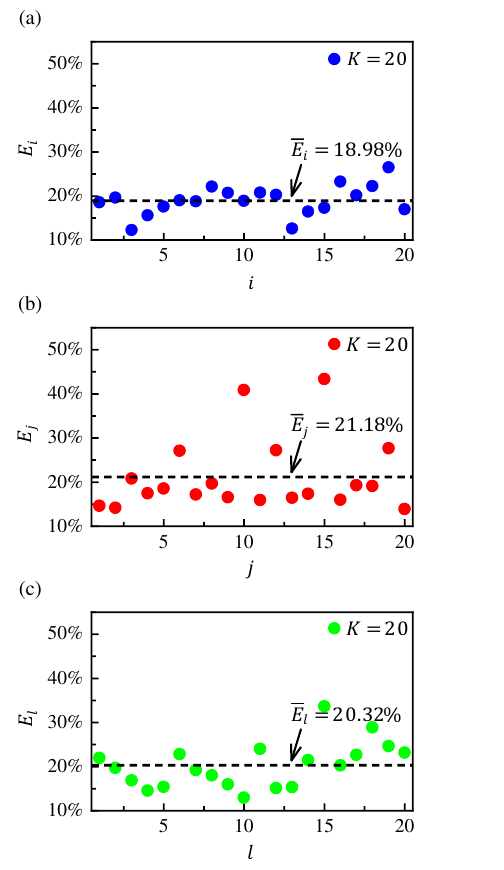}
\caption{\label{App.RE20} Representation errors and the average representation errors for the flow field in subdomains $\Omega_1$, $\Omega_2$, and $\Omega_3$ with clustering number $K=20$. 
}
\end{figure}
The representation errors and the average representation errors 
while $K=10$ and $K=20$ are shown in Fig.~\ref{App.RE10} and Fig.~\ref{App.RE20}.
The lowest average representation error is observed
in subdomain $\Omega_1$ with a clustering number of $K=20$.
As shown in Fig.~\ref{Res}(b) and Fig.~\ref{App.RE20}(a), 
the average estimation error $E(\mathcal{D}_{\mathrm{test}})$ 
at the optimal sensor location $\bm{x}^*_s$ is approximately 19.22$\%$,
unavoidably slightly larger than the average representation error $\bar{E}_i=18.98\%$ obtained with $K=20$.
This suggests a direction for future work, 
namely to significantly reduce the error by enabling interpolation between centroids.

\subsection{Inference matrix}
\label{InferenceMatrix}
\begin{figure*}[hbt!]
\includegraphics[scale=1]{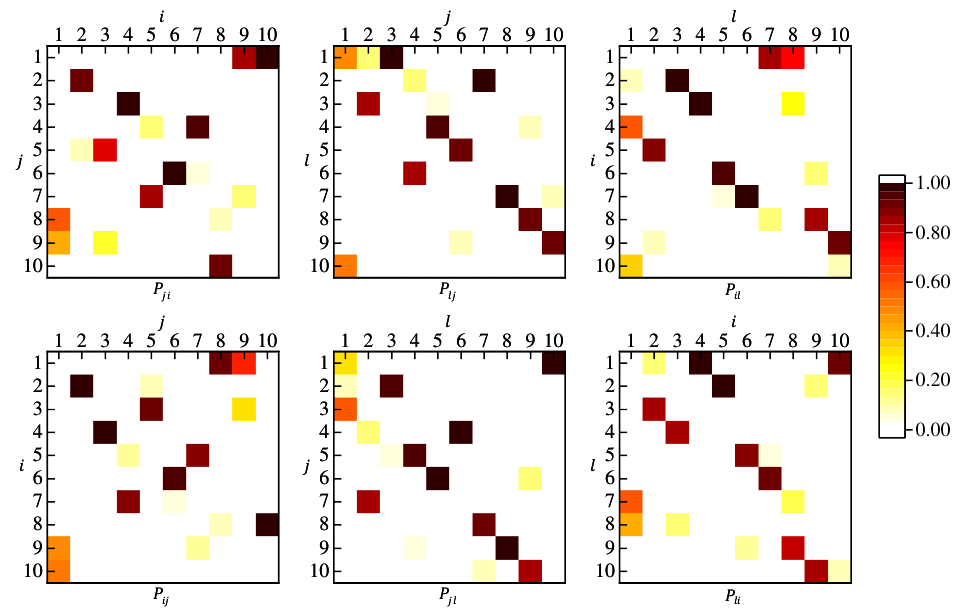}
\caption{\label{App.IM10} Inference matrices with clustering number $K=10$. 
}
\end{figure*}
\begin{figure*}[hbt!]
\includegraphics[scale=1]{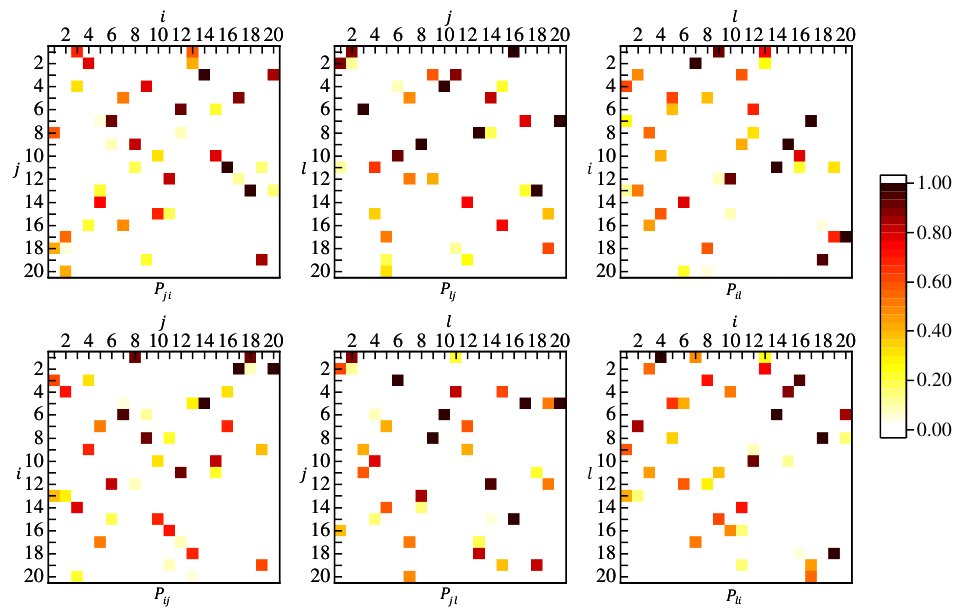}
\caption{\label{App.IM20} Inference matrices with clustering number $K=20$. 
}
\end{figure*}
The inference matrices for $K=10$ and $K=20$ are presented in Fig.~\ref{App.IM10} and Fig.~\ref{App.IM20}, respectively.
With the inference matrices, sensor signals from one subdomain can be used 
to infer the most probable centroid flow state in the other subdomains. 
This capability enables cross-subdomain flow state estimation, 
thereby facilitating efficient estimation of large-scale flow dynamics from localized sensor information.


%% file: S6_App_E.tex
\section{List of symbols}
The symbols are summarized in Table~\ref{tab:qua}.
\begin{table}
\caption{\label{tab:qua} 
List of symbols.}
\begin{ruledtabular}
\begin{tabular}{ll}
Symbols & Variables \\ 
\hline
$\bm{\mu}$ & Operating conditions\\
$\bm{\mu}^m$ & Operating conditions of training set\\
$\bm{\mu}^{M+n}$ & Operating conditions of testing set\\
$U_\infty$ & Wind velocity magnitude\\
$\alpha$ & Wind direction\\
$\beta$ & Drone trajectory position\\
$\mathcal{D}$ & Data set\\
$\mathcal{S}$ & Sensor signal set\\
$\bm{u}$ & Velocity\\
$\hat{\bm{u}}$ & Estimated velocity\\
$\overline{\bm{u}}$ & Mean flow velocity\\
$M$ & Number of snapshots in training set\\
$N$ & Number of snapshots in testing set\\
$\bm{u}^m$ & Training set snapshots\\
$\bm{u}^{m+}$ & Normalized training set snapshots\\
$\bm{u}^{M+n}$ & Testing set snapshots\\
$\hat{\bm{u}}^{M+n}$ & Estimated velocity of testing set snapshots\\
$\overline{\bm{u}}^{M+n}$ & Mean velocity of testing set snapshots\\
$\hat{\bm{U}}_\infty$ & Estimated wind velocity magnitude of testing set snapshots\\
$\Omega_1, \Omega_2, \Omega_3$ & Discretized subdomains\\
$i, j, l$ & Cluster affiliation in $\Omega_1, \Omega_2, \Omega_3$\\
$\bm{c}_{1,i}, \bm{c}_{2,j}, \bm{c}_{3,l}$ & Centroids of $i,j,l$\\
$\mathcal{C}_{1,i}, \mathcal{C}_{2,j}, \mathcal{C}_{3,l}$ & Clusters of $i,j,l$\\
$k_1, k_2, k_3$ & Cluster affiliation function\\
$K$ & Total cluster number of each subdomain\\
$\mathbf{P}$ & Inference matrix\\
$P$ & Conditional probability\\
$E(\mathcal{D}_{\mathrm{train}})$ & Average estimation error of training set\\
$E(\mathcal{D}_{\mathrm{test}})$ & Average estimation error of testing set\\
$E^m, E^{(M+n)}$ & Estimation error of training set and testing set\\
$\bm{x}_d$ & Drone trajectory\\
$\bm{x}_s$ & Random sensor location\\
$\bm{x}^*_s$ & Optimized sensor location\\
$\bm{u}_d$ & Velocity field on drone trajectory \\
$\bm{s}$ & Sensor signals\\
$\bm{s}^{(M+n)}$ & Sensor signals of testing set\\
$\bm{s}^{\bm{c}}$ & Sensor signals of centroids\\
$\bm{s}^{(M+n)^{+}}$ & Normalized sensor signals of testing set\\
$\bm{s}^*$ & Sensor signal from optimized sensor location\\ 
\end{tabular}
\end{ruledtabular}
\end{table}